\documentclass{article}

\usepackage{arxiv}

\usepackage[utf8]{inputenc} 
\usepackage[T1]{fontenc}    
\usepackage{hyperref}       
\usepackage{url}            
\usepackage{booktabs}       
\usepackage{amsfonts}       
\usepackage{nicefrac}       
\usepackage{microtype}      
\usepackage{graphicx}
\graphicspath{ {./images/} }

\title{Livewired Neural Networks:\\
Making Neurons That Fire Together Wire Together
}

\author{
 Thomas Schumacher \\
  Johns Hopkins University\\
  \texttt{tschuma2@jhu.edu}
}

\begin{document}
\maketitle
\begin{abstract}
Until recently, artificial neural networks were typically designed with a fixed network structure. Here, I argue that network structure is highly relevant to function, and therefore neural networks should be livewired (Eagleman 2020): dynamically rewired to reflect relationships between higher order representations of the external environment identified by coincident activations in individual neurons. I discuss how this approach may enable such networks to build compositional world models that operate on symbols and that achieve few-shot learning, capabilities thought by many to be critical to human-level cognition. 
\newline
\newline
Here, I also 1) discuss how such livewired neural networks maximize the information the environment provides to a model, 2) explore evidence indicating that livewiring is implemented in the brain, guided by glial cells, 3) discuss how livewiring may give rise to the associative emergent behaviors of brains, and 4) suggest paths for future research using livewired networks to understand and create human-like reasoning.
\end{abstract}

\section{Introduction}
\subsection{Overcoming Limitations of Neural Networks}
Artificial neural networks have led to rapid advances across a wide variety of tasks, owing to the rapid proliferation of new approaches (see Raghu \& Schmidt, 2020 for an overview). Yet for all these advances, there remain tasks fundamental to human cognition which remain beyond the reach of existing approaches. Many have argued that the core limitation of neural networks is an inability to effectively model relationships between entities in the environment, an approach which is critical for human cognition (for a fuller discussion, see Battaglia et al. 2018).

An alternative statement of this core problem is the binding problem: to acquire a compositional model of the world, models must operate on symbol-like entities. This binding problem has a long history in the fields of neuroscience and cognitive psychology, investigating how the relationships between objects are developed in the mind and the brain. It has long been argued that the inability of neural networks to operate effectively on symbols imposes a fundamental limit on their performance; for a fuller discussion of the binding problem and the power of symbolic methods, see Greff et al. (2020).

Along similar reasoning, I submit that the inability of early neural networks to model relationships between higher order representations of the environment is largely due to the fixed network structure of such networks. Specifically, following on the argument presented by Battaglia et al. (2018), I would argue that the only means for neural networks to form compact representations of relationships in the environment is through the formation of connections between nodes that represent entities in the environment which interact. However, rather than requiring a graph structure of the interactions in the environment be supplied as an input to a model (see Wu et al. 2019 for a discussion of how graph neural networks operate on graph structures), I propose that this structure should instead be generated as part of the learning process to ensure all relationships encountered through inputs from the environment are captured in the network structure.

More specifically, I propose that when a neuron experiences a particularly strong activation, it should seek out coincident strong activations in unconnected neurons and consider forming connections with them based on the values of the loss gradient for the proposed connections to enable the construction of relationship connections analogous to those explicitly provided in graph networks (Wu et al. 2019). I will sketch out below how introducing the ability for neural networks to livewire (dynamically rewire to connect higher-order representations with coincident strong activations) will enable neural networks to efficiently build the complex yet sparse graph structures that most compactly encode the significant interactions between the numerous entities in the external environment.

To more deeply motivate focusing connection growth between pairs of neurons with large activation values, I will illustrate why such an approach should be expected to identify the most promising connections. More specifically, I will discuss how the loss function is a measure of the imperfection of a model and therefore, as the loss gradient for a connection in any neural network is proportional to the activation value of the parent node (Rumelhart et al. 1985), examining pairs of neurons with large activation values will focus connection growth where it will most quickly improve model performance. Moreover, by focusing solely on connections between neurons with strong activations, livewiring this way will overcome the daunting combinatorial challenge of identifying which of the multitude of possible pairs of entities in the environment should be inspected for relationships.

One further advantage offered by this livewired approach is to extend neural networks’ capacity for few-shot learning. In neural networks with a fixed structure, the only means for a network to significantly alter its predictions in response to novel experience is to alter the weights across the entire network, but this approach contradicts the goal of model convergence. However, in a livewired network, we can adjust the learning rate at each connection based on its experience, enabling the network to rapidly improve new connections formed in response to novel inputs, without substantial updates to the broader network that would cause catastrophic forgetting of prior learning. 

This variable learning rate approach will be further motivated by the idea of credibility; the weights on older connections have been fitted using more data, making them more reliable than those for newer connections. This suggests weights for older connections should remain stable to maintain model convergence around prior data, while weights on new connections should update rapidly to accelerate convergence for novel examples.

\subsection{Drawing Further Inspiration From Neuroscience}
Early neuroscience research focused largely on the role of neurons in propagating electrical signals (Ramón y Cajal 1933). Initial attempts to replicate the function of neurons assumed the network structure would be fixed (McCulloch \& Pitts 1943). This simplified model of the brain inspired initial efforts to replicate neural functions using perceptrons (Rosenblatt 1961). 

However, more recent research suggests the brain uses glial cells to not only modulate the existing connections between neurons, but also update the brain’s structure of connectivity, forming long-range connections (Fields et al. 2015). There is evidence that this re-wiring role is essential to enabling human-level cognition; for instance, it has been demonstrated that injecting human glial cells into mouse brains enhances their ability to learn (Han et al. 2013). However, the precise mechanisms by which glial cells help control learning remain an active area of research (Fields et al. 2014, Santello et al. 2019).

I propose that enabling neural networks, artificial or biological, to dynamically rewire in response to activation experience will improve their performance on tasks requiring rapid learning of connections between disparate functional areas. As discussed further below, enabling networks to take neurons that fire together and rewire them together can be viewed as a natural extension of Hebb’s rule, which predicts the strengthening of a connection from a neuron to another that frequently causes the other to activate (Hebb 1949). By way of extension, I argue that pairs of neurons that fire strongly together should often wire together, even if they are only distantly connected.

More to the point, the activations of individual neurons can be thought to represent a signal of a higher-order representation of the environment, which the activations of neurons are by construction. Therefore, strong coincident activations of disparate neurons represent a signal from the environment that the higher-order representations those neurons form are (by Bayesian reasoning applied to an ordered environment) likely to be related in some sense, and learning the relationships between such higher-order representations is most effective if the neurons are wired more closely together. 

I propose that the brain makes use of such connections between higher-order concepts to build efficient models of the environment, which take advantage of the structure of inputs to form compact representations. I will discuss evidence that glial cells are in part responsible for growing new connections and pruning old ones, using chemical signaling to guide this process. I will discuss how glial cells may focus growth on the relationships between higher-order representations of the environment which tend to coincide in that environment, and how this behavior can be replicated in artificial networks.

\section{Background}
To understand more deeply why neural networks with a fixed network structure should be expected to fall short of human-level performance in important respects, it is instructive to more finely contrast the most current understanding of the functioning of the brain, and related observed behavior, with neural networks with a fixed network structure. 

As seen below, the human brain devotes significant resources to updating the structure of its connections, and designs of artificial neural networks have gradually come to better reflect this ability through manual structural interventions in the network, achieving corresponding performance increases. However, including the generation of the structure of neural networks as part of the learning process remains in early stages.

\subsection{Neuroscience}
The ability of and necessity for brains to livewire can be clearly illustrated by considering the brain’s ability to achieve sensory substitution: individuals with sensory deficits, such as blindness or deafness, can learn to sense the world in a roughly equivalent way using different sensory paths, provided they are given access to an alternative sensory input stream that conveys similar information. For instance, blind people can learn to process visual information if electrodes on the tongue stimulate patterns captured by a camera (Bach-Y-Rita \& Kercel 2003). 

Under a livewired interpretation of neural activity, this ability to achieve sensory substitution can be clearly understood: the parts of the brain that form a 3-dimensional reconstruction of the environment (a task critical to survival) will use whatever sensory input that is available to best enable this construction. For sighted individuals, this is most efficiently achieved using sensory data from the visual system. However, sensory substitution demonstrates that the brain can adapt its functional structure to enable other sensory areas to assume the responsibilities normally assigned to another sense, provided it is given access to the right inputs. Moreover, this functional change has been shown to be enabled by significant updates to the connections between brain regions at larger scales (Bach-Y-Rita \& Kercel 2003). See Eagleman (2020) for an in-depth discussion of the power and flexibility of this approach.

To understand how livewiring could be achieved in the brain, it is critical to understand the role of glial cells, which help manage the structural connection of neurons in the brain, among other roles. Glial cells in the brain can be subdivided into separate types: astrocytes, microglia, and oligodendrocytes are most related to the rewiring process. Amongst other functions, astrocytes moderate the formation and destruction of connections, and even growth of new neurons, microglia prune synapses based on activity, and oligodendrocytes can restrict axonal sprouting and are responsible for myelination, which significantly alters the functional structure of the brain, as it determines how well signals transmitted along axons are preserved from noise and how quickly axons conduct action potentials (Fields et al. 2015).

Astrocytes are critical for learning and memory formation (Adamsky et al. 2018). To fulfill this role, astrocytes communicate with neurons (Scemes \& Giaume 2006). More specifically, astrocytes form long processes, parts of the cell which extend far from the cell body, some of which wrap around synapses, giving rise to the term tripartite synapses (Araque et al. 1999, Fields et al. 2015). These processes have been shown under some circumstances to become increasingly motile in the presence of synaptic activation that causes long-term potentiation (a state in which it becomes more likely for a neuron to fire), and to become more motile in response to \(Ca^{2+}\) elevation (Bernardinelli et al. 2014). Astrocytes also guide the growth of neurites, including both axons and dendrites (Fields et al. 2015).

However, one functional aspect of astrocytes that has not yet been fully explained (Bazargani \& Attwell 2016) is that under some circumstances, astrocytes generate intercellular \(Ca^{2+}\) waves in response to neuronal activity (Dani et al. 1992, Barres 2008) which are undirected, and can persist for 5-30 min. (Scemes \& Giaume 2006). 

This is potentially significant in attempting to understand the rules by which the brain grows new connections, as it has been observed that in addition to inducing increased motility of astrocyte processes (Bernardinelli et al. 2014), \(Ca^{2+}\) regulates the growth of axons, the growth of synapses, and the plasticity of existing connections (Zundorf \& Reiser 2011). In particular, the development of axons is guided by growth cones, where axonal migration is regulated by calcium transients (Rosenberg \& Spitzer 2011). The importance of \(Ca^{2+}\) waves to cognition can be demonstrated by the association of their dysfunction with pathologies (Nedergaard et al. 2010, Agulhon et al. 2012, Vardjan et al. 2017).

To further emphasize the potential functional importance of calcium waves in humans, it should also be noted that while astrocytes in the human cortex are larger, with more branching processes than those of other animals such as rodents, astrocytes in humans generate calcium waves with substantially greater propagation speed (Oberheim et al. 2009).

To investigate the significance of astrocytes to learning, one study injected human astrocytes into mice brains, resulting in improved learning across a range of tasks, while also displaying enhanced long-term potentiation. Moreover, the human astrocytes grew to form a large percentage of the glial population of the mice brains, and displayed more rapid calcium wave propagation, further suggesting the importance of this function to learning (Han et al. 2013).

Given the long understood importance of sleep to memory formation (Rasch \& Born 2013), it would be expected that if glial cells were important to learning, they would display significantly different behavior during sleep to enable this learning. Among other suggestive behaviors, astroglial processes retract from synapses during sleep (Bellesi et al. 2015), while microglia eliminate synapses during sleep, and the molecules that signal synapses should be eliminated are increased during sleep (Choudhury et al. 2019). 

It has been observed that human brains evolved to have a higher concentration of oligodendrocytes, which are responsible for the myelination that enables long-range connections (Berto et al. 2019). This suggests that glial cells, and especially the long-range connections they enable, are a critical component of human cognition.

As noted by Bellec et al. (2018), the brain is constantly rewiring, but most especially during learning, further supporting the notion that livewiring is a vital mechanism by which the brain adapts to novel inputs.

In addition to the apparent importance of livewiring to the development of the human brain, it should also be observed that infant brains contain many superfluous synapses which are pruned in early childhood development, and the length of the pruning period is dependent upon the functional area of the brain. For example, pruning in perceptual areas ceases before that for higher cognitive functions (Tierney \& Nelson 2009). This suggests that plasticity is activity-dependent, either by means of some regulatory mechanism which adjusts plasticity in response to experience, or a set plasticity schedule for different brain regions which have evolved to optimize efficiency of the learning process across the lifetime of the brain.

\subsection{Associative Pattern Seeking}
Humans seek causal connections between pairs of events that happen to coincide. This can sometimes lead to the formation of spurious connections between unrelated phenomena. However, it has been argued that humans’ bias towards pattern-seeking poses an evolutionary advantage, as it enables the identification of important relationships in the environment, the value of which outweighs the costs of unfounded conclusions (Foster \& Kokko 2009, Johnson et al. 2013).

While a bias toward pattern seeking behavior may enable more effective causal analysis, it can also pose substantial problems; identifying erroneous causal relationships, a tendency termed apophenia, has long been connected with mental disorders such as schizophrenia (Conrad 1958).

The associative nature of human cognition can most clearly be seen in the phenomenon of priming, by which sensory information that is related to recent experience is more effectively processed. For instance, it has long been observed that when primed with specific words, humans are much better for a short period at recognizing other words that are semantically linked (Foss 1982). A similar effect has also been observed in visual information, where it has been seen to be robust to transformations of the input (Biederman \& Cooper 1992).

\subsection{Artificial Neural Networks}
It has been argued that the key to expanding the representational power of neural networks is to enable them to have a relational inductive bias (Battaglia et al. 2018), in order to enable them to better capture associative pattern seeking.

The power of modifying the structure of networks to achieve more compact relational representations can be demonstrated by the success of approaches that introduced manual structural interventions to encourage such relational representations. For instance, successively higher-level representations achieved through the use of stacked convolutional layers was demonstrated to achieve better results than single convolutions by Inception (Szegedy et al. 2014). Other attempts have also been made to extend the representational power of such manual structural interventions, for instance capsule networks (Sabour et al. 2017) and Squeeze-and-Excitation Networks (Hu et al. 2018). 

Despite the increased representational power manual structural interventions can provide, it should be noted that many current neural network architectures demonstrate an inductive bias towards locality. For instance, it has been observed that deep convolutional networks do not recognize shapes (Baker et al. 2018), small transformations in inputs can cause poor generalization in deep convolutional networks (Azulay \& Weiss 2018), and deep convolutional networks do not take advantage of semantic connections (Devereux et al. 2018).

To incorporate learning the network’s structure as part of the learning process, a number of approaches to rewiring neural networks have been proposed by creating only sparse connections, thus limiting the required number of parameters to learn in the network (Stanley \& Miikkulainen 2002, Bellec et al. 2017, Chen et al. 2018, Mocanu et al. 2018, Liu et al. 2019, Mostafa \& Wang 2019, Billaudelle et al. 2019, Limbacher \& Legenstein 2020). While these approaches differ in their specific choice of how rewiring is performed, they each choose new connections between adjacent layers at random in some sense.

Recently, a new rewiring approach has been proposed that selects new connection growth based on the magnitude of the loss gradient for possible connections (Evci et al. 2019). This approach was largely in response to the fact that most approaches to constructing sparse networks required about as much resources as dense networks to train initially (Zhu \& Gupta 2018; Guo et al. 2016).

To date, it appears no approach to growing new connections across all possible layer pairs has been attempted, likely due to the significant combinatorial cost of examining all such pairings. However, such connections have the potential to greatly improve the performance of networks, as can be demonstrated by the growing success of approaches that rely on such connections.

For instance, highway networks enable higher order representations that prove useful to persist through the network, as they enable the carry-over of previous layers using carry gates, inspired by LSTM’s (Srivastava et al. 2015). Residual learning (ResNets) restricts this idea by eliminating the gating function, enabling all information from the prior layer to carry over to the next layer (He et al. 2016). Densely connected networks further extend the notion of passing activations from prior layers through the network by enabling the passing of inputs and activations across an arbitrary number of layers (Huang et al. 2017). 

Results from using a fractal network structure suggest that effective network depth, rather than the use of residuals, may explain the success of ResNets (Larsson et al. 2017). One possible explanation for the importance of effective network depth is the interpretation of the network as a Markov chain (Tishby \& Zaslavsky 2015), by which we would expect the quality of the signal provided by the loss gradient to degrade as it backpropagates through the network, just as the signal of an activation degrades as it feeds forward. In such cases, effectively shallower networks would enable the loss gradient to be more clearly communicated back through the network.

\subsection{Relational Models}
It has been argued that an important aspect of building compact, powerful world models is to take advantage of compositional rules to combine representations into higher-level abstractions (Bottou 2014). One formulation of this approach that has been proposed is to build models which represent entities and their relationships in the environment using graph structures, with the construction of such graphs being a crucial aspect of the learning process. However, it remains an open question as to how to efficiently induce such a graph structure as part of the learning process (Battaglia et al. 2018). 

Many approaches to using relational representations to exploit structure in the environment have been proposed to construct models that can operate on recognizable symbols. Some of these approaches rely on supplying the structure of the data explicitly as an input to the model, while others attempt to learn the structure from data, though such approaches to date have primarily focused on relatively narrow problem domains. For the sake of brevity, I will list only a subset of the varied approaches to these two related problems.

Neuro-symbolic systems seek to achieve more compact representations of data by structuring representation around symbolic logic (Yi et al. 2018). Deep learning performance can be improved in some problem domains by enabling networks to apply semantic rules, as demonstrated by recent efforts in neuro-symbolic AI (Mao et al. 2019). 

Relation Networks can be used to model sets of objects that interact, though require specifying lists of objects and the interactions that exist between them (Santoro et al. 2017). Other relation networks seek to achieve few-shot learning by learning deep metrics to compare the similarity of query images against few-shot labeled sample images (Sung et al. 2018).

Graph neural networks learn to form representations on graph structures, which are typically provided as input to models. Graph neural networks can be approached using a variety of paradigms, including recurrent neural networks, convolutional neural networks, autoencoders, and spatial-temporal neural networks (Wu et al. 2019).

The successes of these structured approaches to deep learning in their target domains serve to illustrate the importance of network structure to forming useful representations. Indeed, performance of neural networks has been shown to be deeply related to network structure, in particular clustering coefficient and average path length (You et al. 2020).

Yet capturing the full structure of relationships between entities in the environment is a monumental task, and such structure cannot be explicitly encoded, as any intelligent agent operating in the world will constantly encounter novel combinations. Therefore, many approaches to building compact, relational models of the world have focused on integrating the generation of graphs describing the relations in the environment into the learning process. In this way, it is hoped that the entities and their relationships to other entities in the environment will be mapped by the connection structure of such graphs.

For instance, relational neural expectation maximization uses unsupervised visual images to generate models of objects and their interactions. It makes use of a variational auto-encoder operating on a fully connected graph (van Steenkiste et al. 2018).

Graph Transformer Networks learn network structure by searching for new connections along meta-paths (paths along connections across multiple nodes), though require deep representations to learn longer meta-paths, requiring significant resources to form new connections between previously distant representations (Yun et al. 2019).

Yet these operations on graphs must examine pairwise relationships, incurring considerable computational complexity when expanded to the full range of interactions modeled by humans. See the discussion on Overcoming the Curse of Dimensionality below for a discussion of how livewired neural networks can be expected to overcome this combinatorial problem. 

To emphasize the potential of learning richer representations by building dynamic hierarchical models, it should be noted that generating network structure as part of the learning process has shown to be a promising approach to building more effective models. For instance, Liu et al. (2017), NASNet (Zoph et al. 2018), and Xie et al. (2019) incorporated network architecture into the learning process to improve model performance, and Graphite (Grover et al. 2017),  Franceschi et al. (2019), and Alet et al. (2019) introduced learning graph structures into the learning process for graph neural networks. 

However, for environments with a multitude of entities, efficiently prioritizing a small subset of all entities to investigate for potential relations remains an outstanding problem. This is critical, as any approach that attempts to investigate even a small fraction of the possible relations between all entities in the environment will quickly be overwhelmed, necessitating only sparse investigations of the most promising interactions.

\section{Methodology}
While a contemporary view of neuroscience suggests the need to dynamically rewire neural networks in response to activity, which is further hinted by the associative nature of pattern seeking in human behavior, these motivations do not serve as a practical guide for how this may be accomplished in neural networks. 

The method developed below focuses the growth of new connections between pairs of neurons with strong coincident activations. This approach will be motivated by a Bayesian interpretation of neural activity; specifically, by connecting neurons with coinciding improbably high activation values, we assert a conditional relationship likely exists between them. This in turn enables the network to more effectively represent the likelihood of observing this coincidence, with the expectation that a structured environment with entities engaged in sparse relationships will frequently generate such coincidences repeatedly. 

However, first we will examine how models seek to minimize the surprise from their environment. We can then connect this goal to neural networks’ need to maximize the passing of information through the network, which will further motivate the need for neural networks to livewire.

\subsection{Maximizing Environmental Information}
The goal of any learning system is to maximize the intake of information conveyed by the external environment during learning to minimize future environmental information, as any deviation between the predictions of a learning system and the feedback from its environment reflects a defect in its world model. In addition to measuring the level of imperfection of a model, the amount of information the environment conveys to a model also is generally reflective of the amount of resources required by the model to improve its representation to better reflect the environment.

To better understand why the learning problem can be viewed as minimizing future environmental information, consider the definition of Shannon information of a signal x: 

\begin{equation}\label{}
    I_X(x) = -log(p_X(x))  
\end{equation}

which can be interpreted as a measure of surprise, as low probability events have a high information value (Shannon 1948).

For a perfect model of a deterministic world, all future observations could be predicted exactly, in which case \(p_X(x)=1\) for all signals from the environment, so the information provided by the environment to the model would be 

\begin{equation}\label{}
    I_X(x) = -log(p_X(x)) = -log(1) = 0 
\end{equation}

In this case, no new information would be conveyed by the external environment, and no resources would need to be devoted to further learning, as the model would already exactly reflect the environment.

However, for a neural network that assigns a low probability to an outcome that is observed, \(I_X(x) = -log(p_X(x))\) would be quite large, indicating the model is surprised by this event. Thus, the more information the environment communicates to the model, the more the model is surprised, and the worse the model can be said to perform. 

Viewed in this way, a loss function is a measure of the imperfection of a model, as greater deviation between the expected and actual value of an observation implies the model assigns a lower probability to the actual value. Thus, the gradient of a loss function provides a means to measure how quickly model imperfection can be improved upon. Using backpropagation, neural networks can attribute the capacity for model improvements across all connections in the network (Rumelhart et al. 1985). 

Continuing with this approach, the loss gradient at each connection can be interpreted as proportional to the extent to which we can maximize the amount of information learned by the model by altering this connection. Therefore, the network should focus model updates on connections with high loss gradients, as those are the connections that can generate the most rapid improvement.

Internal nodes in a neural network can be interpreted as higher-order representations of the environment by their construction. To improve upon the higher-order representation provided by an internal node, beyond following gradient descent to optimize the weights for connections to the node (Rumelhart et al. 1985), a reasonable next step is to use the loss gradient to help select the growth of new connections (Evci et al. 2019).  

To understand why this approach could be expected to improve the overall performance of the network, consider the Lottery Ticket Hypothesis (Frankle \& Carbin 2018), which hypothesizes that the overall performance of a neural network can be highly sensitive to the initial conditions of the network, with additional connections increasing the likelihood of initializing a “winning ticket” by adding a new degree of freedom. Therefore, adding connections can be expected to improve the higher-order representations nodes form. Moreover, using the loss gradient to guide the growth of new connections will enable the network to focus on changes with the greatest potential for rapid improvement.

Under the interpretation offered by the Lottery Ticket Hypothesis, the poor performance of individual nodes may be due to an unfortunate initialization. Providing greater degrees of freedom to the model in the form of new connections in such cases should be expected to improve the performance of the network; even if redundant information is fed forward in some cases, denser networks are more likely to contain a “winning ticket” than randomly selected sparse networks (see Evci et al. 2019 for empirical results supporting this notion). As learning proceeds, redundant connections should be gradually pruned, leaving only those that are most important for modeling the environment.

\subsection{Relationship Seeking Through Livewiring}
\subsubsection{Motivation}
While growing connections using loss gradients as a guide is a good starting point, as it will focus growth on connections with the greatest opportunity to minimize environmental information, this approach becomes combinatorially impractical if we attempt to expand the growth of new connections to include those across multiple layers, as we must examine the gradient along every node pair in this case. Such a comparison would require a direct signaling mechanism between every pair of neurons in the network, which would be computationally (or metabolically) expensive, requiring \(O(n^2)\) comparisons for networks with n neurons.

For reference, while it is not known precisely how many connections there are per neuron in the brain, estimates are well below that of a fully connected network, with the average number of synapses per neuron estimated to be on the order of thousands, while the total neurons in the human brain are on the order of billions (Dicke \& Roth 2016). Therefore, if a network seeks to find new connections to build across layers or brain regions, it should have some means of identifying promising neuron pairs.

To motivate the need for growing cross-layer connections, recall that it is thought that in the brain, long-range connections are enabled by the myelination process, which helps preserve long-range signals between disparate functional areas for which communication is important (Fields et al. 2015). Given the increased presence of oligodendrocytes responsible for myelination in the human brain (Berto et al. 2019), it would seem the formation of long-range connections within the brain is functionally important, especially for cognitive tasks of which only humans are capable. 

Also, neurons (or hidden nodes in neural networks) can be interpreted as higher order representations of environmental inputs by construction. This interpretation implies that strong coincident activation of neurons, artificial or otherwise, is a signal from the environment that a relationship between those two higher order representations may exist (as discussed further below). Restricting connection formation to only adjacent layers would therefore limit the network’s ability to infer relationships between entities that may be related solely on the basis of the distance between those entities based on the model’s internal representations of those entities.

However, to overcome the combinatorial challenge of forming connections across multiple layers, the approach I will use is to restrict the search for high gradient connections to those between nodes with strong activation values. This will focus growth on pairs of neurons that we have good reason to expect to be related, as their coincident strong activation values suggest their coincident presence in the input. This approach will also limit the space of connections which must be examined for new growth, as it will greatly decrease the number of node pairs required for inspection to a level we can control.

Moreover, recall that the loss gradient for a connection is proportional to the activation of the parent node (Rumelhart 1985), so a strong activation can be expected to lead to a high loss gradient. Thus, limiting the search for connections with high loss gradients to connections between pairs of neurons with strong activations can be expected to find more promising connections.

Forming connections between neurons with strong coincident activations can also be thought of as a natural extension of Hebbian learning; rather than merely strengthening existing connections between neurons that fire together successfully (Hebb 1949), it is rather the case that any pair of neurons that fire strongly together should be wired together in order to better understand the relationship between the higher-order representations of the environment those neurons represent.

Ultimately, livewired neural networks seek to address the problem of identifying the sparse interactions of entities in the environment. In graph neural networks, this is framed as incorporating the graph structure into the learning process (Battaglia et al. 2018). As explored in the discussion on Relational Inference Patterns below, livewired neural networks should induce a relational structure that reflects the structure of the environment itself.

\subsubsection{Bayesian Interpretation of Coincident Activations}
To better motivate why strong coincident activations should suggest a connection between neurons, consider the joint probability of two unusually high activations, \(\sigma_1\) and \(\sigma_2\), for two neurons that are far apart in the network with \(\sigma_1\) occurring in an earlier layer, the probability of which can be expressed as: 

\begin{equation}\label{conditional}
    P(\sigma_1)*P(\sigma_2|\sigma_1) 
\end{equation}             

As the neurons are far disparate in the network, and given the interpretation of the network as a Markov chain (Tishby \& Zaslavsky 2015), the signal of \(\sigma_1\) is likely to degrade as it proceeds through layers. This notion is supported by the improved performance characteristics of models that use connections across layers (Srivastava et al. 2015, He et al. 2016, Huang et al. 2017, Vaswani et al. 2017), which decrease the distortion of the signal offered by nodes in earlier layers as it feeds forward through the network while also decreasing distortion of the loss gradient as it backpropagates through the network.

Given the degradation of signals discussed above, \(\sigma_1\) and \(\sigma_2\) should be expected to be nearly independent within the model given the distance between them in the network, in which case the only representation allowed by the model will be for

\begin{equation}\label{}
    P(\sigma_1)*P(\sigma_2|\sigma_1) \approx P(\sigma_1)*P(\sigma_2) 
\end{equation}

By assumption, \(\sigma_1\) and \(\sigma_2\) are outliers, so this probability is quite small, indicating the model in its existing form is surprised by this input. 

Alternatively, if the two nodes were directly connected, \(\sigma_2\) would explicitly depend on \(\sigma_1\), and thus the model could efficiently learn the weight between them that maximizes their joint probability (\ref{conditional}) for the given observations of the environment. If a causal connection between these higher-order phenomena does exist, forming this connection will enable far more effective inference about the relationship between them.

However, it should also be noted that the short-term memory architecture of the brain, particularly mechanisms that focus activity on items deemed especially salient (Eriksson et al. 2015), may better enable this type of long-range connection without the immediate need for a direct connection. Moreover, given the persistence of signals in working memory (Eriksson et al. 2015), the brain is also able to connect signals that do not coincide exactly, but rather across a window of time. While this ability will not be addressed in the model presented here, attention models (Vaswani et al. 2017, for an overview see Chaudhari 2019) are explicitly designed for this very problem. Pairing these with livewired networks is a promising direction for additional research.

Another interpretation of the rule for focusing on connecting neurons with strong activations is that each neuron that experiences a strong activation is indicating it contains a large quantity of information about the environment. To minimize further environmental information, the model should seek to preserve and exploit this environmental information to form its predictions. 

Recalling the interpretation of the network as a Markov chain (Tishby \& Zaslavsky 2015), the best way for this information to be preserved is to minimize the effective depth of high-information nodes, which will minimize the distortion of this information. This interpretation is consistent with work that has suggested that effective network depth is an important component to building effective models (Larsson et al. 2017).

\subsubsection{Maximizing Mutual Information for a Relational Inductive Bias}
It has been argued that to build learning models which manipulate symbolic representations, models must learn the relational structure of the world by focusing on learning the relationships between entities that are linked, adopting a relational inductive bias (Battaglia et al. 2018, Greff et al. 2020).

An alternative statement of this approach is that there exist very sparse, specific dependence relationships between entities in the environment. Adopting a graph neural network approach to representing dependencies between entities (as suggested by Battaglia et al. 2018), such dependence relationships could be modeled by adding connections between related entities to enable the passing of information between related entities.

In non-graphical neural networks, the features which individual nodes represent can still be linked to recognizable entities; output neurons are explicitly linked to such entities, while nodes that are heavily weighted in identifying specific outputs can also be said to be related to specific features of that entity. Indeed, it has been demonstrated that neural networks already form lower-level constructions as instrumental features that can be used to identify the higher-level representations that they compose (Wang et al. 2017).

For entities that are related, then, there should be information shared between the features that are used to identify each distinct entity, as the dependence relationship between related entities implies that the presence of one entity should also provide evidence for its related entities.

The most pressing question that then arises is how should the relational structure between entities be constructed; in particular, when should relationships between entities be formed, as represented by connections between their respective network regions? 

Informally, when particular entities in the environment are strongly identified, some of their features must take on strong activations indicating the presence of that entity. For related entities, then, one would expect the features which indicate the presence of those related entities to be strongly activated simultaneously. 

To express this more formally, we can use mutual information:

\begin{equation}\label{}
    I(X;Y) = \sum_{y \in Y}\sum_{x \in X} p_{(X,Y)}(x,y)log\Bigg(\frac{p_{(X,Y)}(x,y)}{p_{X}(x)p_{Y}(y)}\Bigg)
\end{equation}

In the case of related entities, their relevant features will tend to assume strong coincident activation values, each of which have a low probability of occurring. If those feature detectors remain only distantly connected in the network structure of a neural network, then those activations will have nearly independent representations in the network, as the interpretation of neural networks as a Markov chain (Tishby \& Zaslavsky 2015) suggests that distance between connected features will degrade the flow of any information between them.

This initial independence of features of related entities will restrict the mutual information between related entities to values near zero. This is undesirable, as any relationship between entities dictates that the joint distribution of the presence of both entities should be substantially greater than the product of their independent probabilities, meaning their mutual information would be well above zero.

By adding connections between features with strong activation values, we introduce a means for the model to pass information between related entities, thereby allowing the model to learn to represent a joint distribution on the related features. In this way, the connection structure of the model will come to reflect the dependence relationships encountered in the environment. 

This approach relies on the assumption that strong coincidental activations are due to genuine dependence relationships between the corresponding entities. While this assumption is reasonable in a highly ordered, relational environment, it must also be acknowledged that some coincidental activations will be merely the result of noise. 

However, identifying spurious patterns is something humans experience as well (Foster \& Kokko 2009), suggesting that there may be no universal solution to this problem for association-based learning agents. One potential advantage of artificial agents in comparison is that their learning rates can be manipulated to encourage them to keep an open mind.

\subsection{Making Neurons that Fire Together Wire Together}
Following from the discussion above, we can see that neurons with strong activation values should be prioritized for inspection for connection formation. However, before exploring how this will be accomplished, we should more firmly establish which neuron pairs should be considered for new connection growth.

First, preventing the formation of connections within a layer may be advisable; while we could define an ordering of nodes within a layer to ensure cycles (a path of connections starting and ending at the same node) do not occur (as they are not allowed when using backpropagation), connections within layers will disproportionately form if allowed to do so due to correlation caused by shared inputs, which weakens the Bayesian justification of such connections. 

If intra-layer connections were allowed to form, they would likely dominate the process due to the correlation between nodes in a layer, and such connections would be largely redundant due to their shared inputs. This in turn may lead to the co-adaptation of features in the same layer, a problem which requires regularization to prevent (Hinton et al. 2012). While pruning could be used to eliminate the most redundant connections, this would likely result in the model churning between forming and pruning such redundant connections, so preventing the formation of such connections seems a safe policy to prevent this from occurring.

More generally, it could be argued that connection formation between distant layers should be favored over that for closer layers, as closer layers are more likely to contain correlated nodes that should be expected to activate together due to existing connections. In such cases, the coincidence of strong activations can be partially explained by the existing network structure, weakening the Bayesian justification for forming new connections. Once again, this could lead to further churn in the model through growing and subsequently pruning redundant connections in adjacent layers.

To combat this problem, one could adjust the priority rating for proposed connections by a factor to account for the layers between the connections (or some other connection metric, such as the length of the shortest path connecting the two nodes if such a path exists, which may be a better proxy for the level of dependence between the two nodes). One could also prevent the formation of connections between nearer layers by imposing further connection formation restrictions beyond the same layer. 

However, as suggested by Evci et al. (2019), forming potentially redundant connections between adjacent layers can help knock networks out of local optima (or into a “winning lottery ticket”), so restricting such connections entirely is not necessarily a favorable approach, as it would preclude connections that have been empirically suggested to be useful.

Moreover, generating connections between more distant layers disproportionately would tend to compress the effective depth of the network over time. This could decrease the expressive power of the network, as deeper networks have been seen to be more expressive (Eldan \& Shamir 2016). For this reason, it is unclear whether there is value in any adjustments to prefer long-range connections over short-range ones, though this issue is deserving of further investigation.

Now, to identify which neurons have activation values that are unusually strong, there are at least a couple of approaches we could employ. First, we could record each neuron’s typical firing behavior as a baseline comparison, and compare activation values to the baseline for each neuron to determine how relatively strong the values are. While an elaboration of this approach may enable artificial neurons to better approximate the complex behavior of individual neurons, it would also impose additional system memory costs.

Alternatively, if we use batch normalization in training our network, we can manipulate the distribution of activation values in each layer of the network. In this case, we can simply read off the normalized activation values to determine the relative strength of each activation in the network. 

Given the demonstrated regularization advantages provided by batch normalization (Ioffe \& Szegedy 2015), and its usefulness in determining the relative strength of activations without imposing additional memory costs, this approach appears most efficient. However, the benefits of this approach in comparison to the more elaborate individual neuron approach remains an open question.

With these rules in place, one reasonable approach to implementing livewiring is to maintain a priority queue of high activation nodes through the feedforward process, with only a bounded length. Then, once the output has been generated, the loss gradient for each pairwise connection between the nodes in the priority queue will be inspected (subject to the within-layer constraint), and those connections with the highest loss gradients will be formed. 

Stochastic gradient descent (Bottou 1998) can then be applied to the same inputs to learn the weights of the new connections. One promising approach to consider is to interleave connection formation and stochastic gradient descent. This approach is especially interesting for its potential to further explicate humans’ tendency towards curiosity, as discussed in Extensions of Few-Shot Learning below.

Examining this proposed process closer, we see that there are two further hyperparameters which must be considered: 1) the length bound for the priority queue, which determines how many high activation nodes to track, and 2) the number of new connections to grow. 

Given the subsequent loss gradient comparison, the length bound is likely best set to the largest value that is computationally feasible, though setting the bound very high could cause obstacles to learning in the case of vanishing or exploding gradients, given the reliance of connection formation on the loss gradient. While I will not specifically address this potential problem here, it is an interesting open question for further investigation.

However, in the case of the new connection formation rate, this hyperparameter appears strongly related to the overall plasticity of the network, in which sense it can be compared to the learning rate. For learning rates, it has been shown that a cyclic approach in which the learning rate initially increases, followed by a period of learning rate decrease, enables faster convergence (Smith 2017, Smith \& Topin 2017). Therefore, it seems advisable to use a schedule for connection formation which initially encourages a high and growing connection formation rate, but which then decreases connection formation over time. 

To maintain the sparsity of the network, for each connection which is formed, a proportional number of connections in the network should be dropped, starting with those with the smallest absolute weights, as smaller absolute weights indicate weaker relationships between features. The ratio of pruned connections to new connections should be adjusted over time consistent with the cyclic approach described. This would be consistent with the observation that brains tend toward greater sparsity as they age (Tierney \& Nelson 2009), and that connections appear to grow denser in early development, the timing of which depends on the brain region involved (Huttenlocher \& Dabholkar 1997).

\subsection{Network Initialization and Regularization}
Brain development in infants (Tierney \& Nelson 2009) and the Lottery Ticket Hypothesis (Frankle \& Carbin 2018) both suggest there is an advantage to initializing denser network structures than the desired ultimate model, using pruning to make the model sparser as genuine connections are discovered.

Empirical evidence from FractalNet suggests initializing network structure to minimize the effective depth of all layers of the network improves model performance, so following the approach of FractalNet, each layer should be initialized to connect to every later layer with a sparsity hyperparameter that shrinks exponentially across each layer (Larsson et al. 2017). 

This approach is loosely supported by evidence from macaque visual cortices indicating connection density decreases exponentially with physical distance (Markov et al. 2014), and evidence that such exponential power scaling applies to human brain activity as well, though that scaling factor appears to depend upon brain region (Tomasi et al. 2017).
More explicitly, the initial connectivity between each layer should be characterized by the following sparsity measure:

\begin{equation}\label{}
    sparsity = sparsityHyperparameter * e^{branchingFactor*layerDifference}
\end{equation}

This structure is intended to minimize effective network depth while constraining the overall density of the network, serving as a generalization of the structure described in Larsson et al. (2017).

Dropout (Hinton et al. 2012) should be used, as constructing redundant representations to avoid problems of co-adaptation of features appears likely to be as beneficial for livewired networks as it has proven in other contexts. The separation of redundant features enabled by dropout (Hinton et al. 2012) may be especially important for enabling overloaded entities to split into separate entities based on distinct associations, as discussed further in Relational Inference Patterns below.

As mentioned above, batch normalization should also be used to encourage stationarity of activation distributions (Ioffe \& Szegedy 2015), and to measure the relative strength of activations. This approach is widespread, as it is important to maintain the same distribution for inputs to nodes to ensure prior learning remains valid, which may not be the case if activation distributions were allowed to shift. This is especially important for livewired networks, as it would not be practical to measure the relative strength of an activation with a shifting activation distribution.

\subsection{Enabling Few-Shot Learning}
One of the core questions about learning in humans is how we are able to generalize learning from small sample sizes, while maintaining a stable model of the world not subject to forgetting (Sung et al. 2018). To begin to address this problem in livewired networks, I propose a generalization of the conventional learning rate schedule approach whereby the learning rate is decreased over time (Bengio 2012); rather than adopt a fixed learning rate schedule across all connections, I propose instead that each connection should adopt a separate learning rate that is initialized to a high value upon the formation of the connection, and which is gradually decreased.

To begin to justify this approach, I would argue that 1) more recently formed connections should remain plastic so that spurious connections based on few data quickly learn more appropriate values than their random initializations, and 2) older connections should grow less plastic to facilitate model convergence and stability. By focusing learning on new connections, this approach should minimize changes to the broader model, thereby avoiding the problem of catastrophic forgetting (Kirkpatrick et al. 2017). For a more in depth discussion of this approach, see the discussion on Credible Credit Assignment.

The use of momentum in updating model weights, whereby updates at the current learning step are weighted between the loss gradient and updates from prior learning steps (Qian 1999), has been empirically shown to minimize model churn and thereby accelerate learning. This benefit appears likely to persist across all learning based on some form of gradient descent, as current learning that contradicts prior learning should be largely ignored, while updates that persist across separate instances should be magnified, which can be achieved using momentum to cancel out contradictory weight updates while amplifying congruous weight changes. Therefore, using momentum appears advisable, with the only modification being that connections that have been eliminated should not be updated.

\section{Discussion}
\subsection{Overcoming the Curse of Dimensionality}
In learning systems, the curse of dimensionality is used to refer to problems with high dimensional input, which poses a challenge to the ability for most models to generate meaningful predictions without large volumes of data or some means to represent the data more compactly. For relational models of the world, such as language, we encounter the curse of dimensionality when considering the full range of all possible interactions between each pair of entities which may interact. There are far too many such possible interactions to be able to model effectively, so for learning to be possible, only a small subset of all possible interactions must be considered.

The key to overcoming the curse of dimensionality of relational models in environments with many entities is to only model sparse interactions. Some existing approaches enable human experts to explicitly map the relationships that exist in the environment. For instance, probabilistic graphical models vastly simplify the learning of a probability distribution over a large group of random variables by making explicit dependence assumptions about the relationships between entities (see Koller \& Friedman 2009 for details). 

However, to this point, it has been an open question as to how the generation of such structure in the case of neural networks may be fully integrated into the learning process (see Battaglia et al. 2018 for a discussion). Any solution must overcome the curse of dimensionality by identifying some mechanism by which proposed interactions can be prioritized for inspection, as attempting to reason over the full set of all possible interactions between all pairs of entities will be computationally infeasible in an environment with numerous distinct entities.

By forming connections between nodes which form higher-order representations of entities that coincide in the environment of a given input, livewiring will overcome the curse of dimensionality by focusing on learning the relationship between entities that are currently present in the environment. Moreover, the enforced sparsity of the network will ensure that the massive combinatorial requirements of a fully connected model will be greatly reduced, focusing learning only on relations between entities that are related in the environment.

\subsection{Relational Inference Patterns}
For an example of how a compact relational model can be formed that relies on sparse interactions that can prioritize specific connections between entities, it is helpful to consider language. Language can be viewed as a model of the world, one that exploits dependencies within the environment to form compact representations that rely on relations between entities. 

The relational structure of language can be demonstrated by the success of hidden Markov models in modeling language, most often applied to speech data (Juang \& Rabiner 1990). Without an inherently association-based structure, language could not be represented so successfully with such a simplistic association-based transition model.

To understand how livewired neural networks will enable the construction of such relational models with a similar ability to exploit the same types of dependencies as language, I will examine a few types of connection patterns that are essential to the relational structure of language, and which should be learned by a livewired network structure.

\begin{figure}
  \centering
  \includegraphics{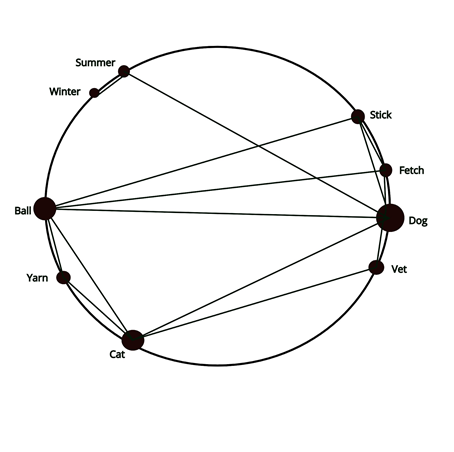}
  \caption{Concept surface on a small group of words.}
  \label{fig:fig1}
\end{figure}

To guide this discussion, it will be instructive to visualize the relational structure of language. To do so, it helps to consider “concept surfaces”, which represent relations between entities as connections between nodes that are related. I use the term “concept surface” and the depiction of Figure \ref{fig:fig1} to loosely suggest that such internal connections between nodes on an outer surface may be reflected in the connection structure of the cortex.

In any relational approach to modeling the interactions of entities, there are a variety of inference patterns that are essential. While it is beyond the scope of this work to prove here the full range of inference rules which are required for a human-level reasoning system, I will at least illustrate how livewired neural networks can be expected to be able to generate structures required for several key inference patterns. For a clearly interpretable explanation of such patterns, it is helpful to reflect on structure in language.

The simplest type of interaction which relational models should form are direct associations between entities. Such interactions can be thought of as the sparse connections between entities in a concept surface, as depicted in Figure \ref{fig:fig1}. The presence of such direct connections in the human brain can be demonstrated by the phenomenon of semantic priming, whereby words that are semantically linked to recently encountered words are more easily identified (Foss 1982).

To understand how livewired neural networks can be expected to induce such an associational structure, note first that in any neural network that has been tasked with a supervised learning problem of identifying specific entities, there are specific nodes which are associated with specific entities. In particular, the output nodes for specific entities are explicitly meant to represent that entity, but also hidden nodes which are heavily weighted in that entity node can also be said to represent that entity to a lesser degree. Thus, it can be seen that in any neural network that can identify entities, there are specific nodes in that network which are at least loosely linked to a particular entity.

For associations to be drawn between related entities, nodes that roughly represent those related entities should be connected, as they are in the underlying concept surface. In a livewired neural network, if two entities coincide, then some subset of the nodes linked to each entity should be expected to fire strongly, enabling the simultaneous recognition of those entities. When this happens, the nodes that loosely represent these two entities will be prioritized for new connections given their strong activations. 

This will enable livewired neural networks to induce an associational structure that is homomorphic to the underlying concept surface. Put differently, the network structure of livewired neural networks will come to reflect the associational structure of the environment about which they learn, enabling them to form compact relational models. If we assume such entities are represented in the brain by specific groups of neurons in the neocortex, which has been observed previously (Quiroga et al. 2005), a livewired model of the brain would predict that regions that represent related entities would be connected, enabling the structure of the brain to come to reflect the structure of the environment it seeks to represent.

One possible objection to the need for adaptive network structure is that networks with a fixed structure could simply pass forward representations from a lower layer, and thus replicate the same structure as that generated by livewiring. However, recall the interpretation of a neural network as a Markov chain (Tishby \& Zaslavsky 2015), by which we should expect the signal of a feature from an earlier layer to degrade as it passes through the network. 

Indeed, the power of preserving such representations from earlier layers has been demonstrated by the improved performance of models that use connections across layers (Srivastava et al. 2015, He et al. 2016, Huang et al. 2017, Vaswani et al. 2017). Thus, while fixed networks should be capable of forming representations identical to livewired networks in theory, there is no guarantee as to the efficiency of such a process, and in practice fixed networks without skip connections have been shown to be far less efficient in accomplishing this.

Alternatively, one may object that a sufficiently dense network that incorporates connections across all layer pairs (Huang et al. 2017) could come to learn the same representation as a livewired network by allowing all possible connections to form. However, the computational cost of such an approach would be far too prohibitive, and the redundant connections would introduce substantial noise, requiring far more data to learn effectively. This combinatorial problem can only be overcome by models which incorporate network structure into the learning process, generating a far sparser network than the fully connected one.

By a similar argument as for associational structure, we can also demonstrate how livewired neural networks will enable hierarchical representations that exploit compositional structure. In particular, in a supervised setting, taxonomic structure will be generated by a similar process as that for associations, as nodes linked to a specific type and other nodes linked to its broader class will tend to both be activated, causing connections between them to be inspected for formation. Over time, this will lead higher-order entities (in taxonomic terms) to form more connections to lower level items, inducing the same structure as any hierarchy.

Moreover, for the components of a specific type that distinguish it from a broader class, the distinguishing components will be most active when the specific type is active, enabling livewired networks to form connections between the distinct features and the more specific class. While this may also generate connections between the distinct features and the broader class, they should be most present for the specific class, enabling the model to distinguish the specific class from other members of the broader class.

One important pattern in understanding natural language is the overloading of some words with multiple distinct meanings which depend on context. Framing this in terms of a concept surface, there can be multiple nodes with the same label, but which form different associations. To enable this behavior, dropout should likely be helpful.

To understand how dropout could enable effective models of overloaded representations, recall that dropout solves the problem of co-adaptation of feature detectors, ensuring neural networks do not converge to redundant pathways, making them less prone to overfitting (Hinton et al. 2012). In the context of livewired neural networks, this also presents a means by which a single word can be split into distinct representations. 

Dropout helps create distinct pathways by which an entity can be identified (Hinton et al. 2012). Over time, it seems reasonable that this would lead words with multiple distinct associational structures to split these pathways according to the neighborhood of associations they belong to. An important open empirical question is how best to induce this subtlety-seeking behavior without forming irrelevant distinctions.

Another significant pattern in natural language understanding is compositional concepts that combine distinct entities into a new entity with associations unique to the combination. For instance, “neural” is associated with subjects loosely tied to the brain and “network” is associated with any graph structure of nodes with connections between them, but “neural network” forms its own unique associations with entities such as “deep learning” or “backpropagation.”

To understand how livewired neural networks can help form such compositional concepts, consider what will happen to the child nodes of two connected entities in a livewired network: they will be strongly correlated with the two connected entities, as their two values will flow directly to their child nodes. This will lead to the formation of compositional child nodes which can combine two entities into a new, distinct entity, which can in turn form its own unique associations. This may be expected to give rise to compositional concepts with unique associations. Alternative livewiring rules may also extend this capability, as discussed further in Alternative Livewired Approaches below.

Two related inference patterns which livewiring may also enable are inductive inference (reasoning based on evidence) and deductive inference (deducing consequences from assumed premises). To understand how livewiring could enable inductive inference, note first that we can think of induction as finding a latent representation which relates to multiple observed phenomena. Also, note that we will enforce sparse associations within our model to ensure only entities which are related are connected. From this, it follows that in the presence of multiple entities with a strong association with a latent entity being sought, a relational model can be expected to efficiently identify the entity at the intersection of the given entities. 

For instance, if nodes representing “stick,” “fetch,” and “vet” are all strongly active, the activations of nodes representing “dog” will tend to be strong as well, even absent direct stimuli. In this way, inductive inference can be thought of as a direct consequence of associational priming.

Similarly, for deductive inference, chains of reasoning always follow paths of associated representations, so the relational structure which livewired networks generate should better enable deductive inference. However, the question of how to prioritize such operations on symbols is an interesting open question.

One further representation pattern livewiring may enable is more effective representations of concepts that combine entities, such as sentences. To understand why, note first that sentences can be interpreted as hypergraphs on multiple words (hypergraphs are a generalization of graphs that allow for relations between more than two nodes at once). The compositional relations formed by child nodes may enable the construction of such hypergraphs on combinations of higher-order representations, enabling more direct inference on such compositional higher-order entities.

While such constructive methods for combining elements to form higher-level representations are promising, it is also natural to ask whether livewired neural networks will be able to form deconstructive representations which extract important components to distinguish between higher-level concepts. This is particularly important for the formation of efficient, compact lower-level representations that can be composed to form connections to other higher-level concepts, which are the only level where supervised feedback can be provided to networks. 

It has been demonstrated that neural networks with a fixed network structure loosely form such lower-level constructions as instrumental features that can be used to identify the higher-level representations that they compose (Wang et al. 2017). This suggests livewired networks should also be capable of forming such lower level representations, enabling them to deconstruct higher-level concepts provided by a supervision process into lower-level components. Moreover, livewiring should ensure that the signal offered by such low-level components is fed forward more efficiently to the higher-order representations of the network to which they are related.

From these compositional reasoning patterns livewiring may enable, we begin to see how livewiring could give rise to symbolic operations in livewired neural networks. Namely, the network structure of livewired neural networks should be expected to come to reflect the relational structure of the higher-order representations it seeks to model, enabling the formation of compact representations of the environment.

\subsection{Credible Credit Assignment}
I have introduced the idea that the plasticity of new connections should be greater than that of older connections, as newer connections have been fit on fewer examples. However, while introducing a variable learning rate across different connections based on their age and experience appears a reasonable approach to achieving few-shot learning without sacrificing model stability, a more sound theoretical justification will prove instructive, and may suggest further modifications to the rules for modifying network plasticity to improve model performance on few-shot learning tasks.

One of the core questions in neural networks is how to assign blame for poor predictions to the weights placed on particular connections, a problem known as credit assignment (Minsky 1963). While backpropagation (Rumelhart et al. 1985) attempts to solve this problem when using a fixed network structure, it is natural to question whether the same approach is appropriate in livewired networks.

To explore this question, it is helpful to first consider the optimal approach to using backpropagation to update weights in fixed network structures. Empirically, a cyclical learning rate which first increases then decreases has been shown to enable faster convergence (Smith 2017, Smith \& Topin 2017).

To understand why this cyclic approach should be expected to enable faster convergence, consider first the later period where the learning rate decreases. In this case, we can say that it is reasonable for the plasticity of the model to decrease over time, as the initial representation is likely to be far from optimal given the random initialization of weights in a network, while later in the learning process the weights should be closer to an optimum, suggesting smaller updates should be made.

However, this does not explain why initially setting the learning rate lower than its maximum value and allowing it to increase initially should enable faster convergence. To understand this point, consider 1) the premise that the high-dimensional space of possible weight combinations for neural networks will be dominated by poor representations, with only a very small subset containing useful combinations, and 2) the weights of neural networks are initially set to randomly chosen values.

Assuming this first premise is true, we can begin to understand in a heuristic sense why a learning rate that is initially very high would not be productive. Namely, a high learning rate will cause large updates to the weights at each step, corresponding to a large movement in the high-dimensional weight space. However, given the initially random weight values and the very high dimension of the space involved, the loss gradient is not likely to point directly towards an optimum. Therefore, by setting the learning rate too high initially, the model will churn between approximately random initializations, impeding the model’s ability to organize the weights in a meaningful way. 

As the weights begin to organize, though, the loss gradient will come to better reflect a direction for weights to change that will actually be productive, explaining why the learning rate should increase initially to enable faster convergence, but then begin decreasing once the weights have organized. 

To emphasize the value of this approach, it is natural to compare the initial increase in plasticity followed by a gradual decrease to the connection growth strategy of the brain over the course of development. In particular, following an initial growth stage where many new connections are formed, connections are pruned as brains age, moving the brain toward greater sparsity (Tierney \& Nelson 2009). 

This approach of initial growth followed by substantial pruning is also consistent with the Lottery Ticket Hypothesis (Frankle \& Carbin 2018), which would suggest such an approach would help efficiently discover a valuable representation while minimizing subsequent metabolic cost by eliminating extraneous connections. Using this approach as a guide, it is reasonable to conclude that this same cyclical connection growth strategy could be valuable for livewired networks as well, with more connections being formed than destroyed initially, followed by a period of connection pruning.

However, to return to the question of how to justify a variable learning rate, and to clarify how such a strategy should be implemented, we should seek to better understand the reasoning behind plasticity adjustments over time. To this end, credibility theory can help to clarify this important question.

Credibility theory is typically used to develop estimates for random variables when there are two data sources: a large data set made up of a mixture of random variables, of which the random variable of interest is merely a subset, and a smaller, more specific data set that considers only that subset. 

For instance, when predicting mortality rates for their customers, insurers rely on data for the broader population (which may not reflect the risk factors specific to their customers) and the mortality data for their customers (which should better reflect future experience, though often does not provide sufficient data, hence the need for applying credibility). To estimate the random variable, a credibility factor which depends on the size of the more specific data set is applied to each dataset to estimate the moments of the random variable of interest, with larger specific datasets being given greater credibility (Longley-Cook 1962).

Using this notion of credibility as a guide, it becomes quite clear why a gradually decreasing learning rate is a valuable approach to achieve convergence; as the model continues to learn, its weights have been adjusted by ever increasing amounts of data. Therefore, the relative credibility of new observations should be increasingly discounted over time, as the existing weights of the network are based on progressively larger amounts of data.

However, in considering the question of how backpropagation should be applied in livewired neural networks, this credibility-driven view of network plasticity strongly suggests that a new approach to learning may be needed. In particular, for older connections, the weights will continue to be based on progressively large data volumes, suggesting the same learning rate schedule approach should be appropriate. 

Alternatively, for newer connections, the weights should be considered to have low credibility, as they have not been trained on much data. Under this credibility-driven view, for new connections, the learning rate should be initially set high and set to decrease as the weight becomes more credible with more training data. While it may be expected that such an approach could potentially drive models to substantially faster convergence, particularly for novel inputs, this is an important empirical question that deserves further investigation.

This approach may also better enable networks to avoid the problem of catastrophic forgetting, whereby prior learning in networks is often discarded using more conventional learning approaches, though there have been some efforts to address this problem (for example, see Kirkpatrick et al., 2017). More specifically, livewired networks which apply credibility-based credit assignment may leave existing learning largely intact while enabling new learning based on new connections for novel inputs.

One further question this credibility-driven view of weight plasticity raises is whether a set learning rate schedule is appropriate. In particular, consider the learning rate for connections which have large loss gradient values, which suggests the connection is performing poorly on those connections. In this case, the credibility assigned to the weight on this connection should likely decrease, as it appears not to reflect the external environment. 

In such cases, the learning rate for such connections should likely be increased, which will increase the plasticity of the network regions which are performing poorly. Viewing the connection formation rate as another hyperparameter influencing the plasticity of the network, it may also be advisable to favor connection growth in network regions with large loss gradients. How best to accomplish these changes is an interesting open question.

\subsection{Livewired View of Neuroscience}
Beyond the potential for livewiring to improve neural networks’ ability to form compact relational models, it can also offer a compelling explanation for a wide range of observations about the brain whose functional purpose has been poorly understood.

For instance, livewiring could explain the undirected \(Ca^{2+}\) waves in astrocytes (Scemes \& Giaume 2006). Astrocytes have been demonstrated to be responsible for guiding the growth of new connections between neurons, relying upon intracellular calcium oscillations to fulfill this function at least in some cases (Kanemaru et al. 2007).  The uniform propagation of extracellular calcium waves in astrocytes (Scemes \& Giaume 2006) could thus be interpreted as focusing growth between neurons whose activations happen to coincide.

Under this view, a neuron that experiences unusually high frequency activation would trigger a calcium wave in the local astrocyte population, which indicates to the astrocyte network that that neuron is open to forming new connections with any other surrounding neurons with strong activations. When a coincident calcium wave occurs in a nearby area, then, connection growth should be expected between the neurons that triggered the coincident waves. In this way, new connection growth would be guided along neural pathways that have experienced recent strong activation. This interpretation would also explain the persistence of these \(Ca^{2+}\) waves across time scales of up to 30 min. (Scemes \& Giaume 2006); to enable the identification of patterns across time, particularly involving stimuli that evoke strong activations.

One poorly understood observation that may be explained by livewiring is that astrocytes are tiled, forming clearly demarcated areas of influence that do not overlap with those of other astrocytes. Under a livewired interpretation of learning in which astrocytes are heavily localized, the brain can clearly identify the specific areas which have given rise to calcium oscillations and consequent morphological changes, including increased motility of astrocyte processes and growth of neurites (Fields et al. 2015). 

If astrocyte processes were permitted to extend to multiple regions, the origins of the calcium oscillations they generate could not be as clearly identified, which would interfere with the rewiring of neurons that are firing together. In this way, the tiling of astrocytes can be thought of as a coordinate system by which the brain can identify the source of calcium transients caused by synaptic activation.

One additional observation that can be explained by a livewired interpretation of learning is that neurogenesis is largely limited to the hippocampus and striatum in human adults (Bergmann et al. 2015). Under a livewired view of the brain, this would be expected for humans, as the striatum is (among other functions) thought to be responsible for learning habitual motor tasks (Graybiel \& Grafton 2015), a task which can vary greatly across humans’ lives due to tool use. Similarly, the hippocampus is strongly associated with learning (van Strien et al. 2009), so a livewired model of the brain would anticipate the hippocampus requiring more substantial updates to its network structure over time than other brain regions.

One potential advantage of artificial neural networks in comparison to how brains appear to handle this rewiring process is that artificial neural networks will not be as constrained by the metabolic constraints experienced by brains in finding paired neurons to rewire together. More specifically, in brains, calcium waves remain localized (Scemes \& Giaume 2006), as flooding the entire brain with a calcium wave for a single neuron or small group of neurons would be quite metabolically expensive, as would growing new connections to connect far disparate brain regions, especially as those new connections may not prove useful. 

However, in artificial neural networks, any neuron that experiences a strong activation could simply be added to a priority queue according to the strength of its activation, and all pairwise connections between neurons in the queue could be inspected for potential growth, regardless of their position in the network. This could enable livewired artificial neural networks to be even more effective at identifying long-range relationships than brains. 

Alternatively, there may be some benefits to the localized nature of connection growth in brains. To understand why, it is critical to first articulate how entities with far disparate representations can come to be linked in a brain constrained by localized connection growth. In particular, it appears that the brain makes use of centralized hubs which play a critical role in memory and learning.

The two brain structures which appear most related to this role as centralized hub are the prefrontal cortex and the hippocampus. The prefrontal cortex plays an important role in focusing attention on particular entities, and is thus thought by some to be responsible for working memory, though there is also evidence to suggest that working memory is at least partially a distributed process with the prefrontal cortex merely pointing to representations elsewhere in the brain (Lara \& Wallis 2015). 

Meanwhile, the hippocampus is thought to be responsible for encoding, consolidating, and retrieving declarative memories (van Strien et al. 2009), and to play a significant role in imagination, combining existing representations in novel ways (Hassabis et al. 2007).
The prefrontal cortex and hippocampus are also thought to be modulated by brain regions responsible for maintaining emotional states, such as the amygdala (Salzman \& Fusi 2010, Tyng et al. 2017).

Circling back to the question of how localized connection growth could benefit the brain, the connection of the prefrontal cortex and hippocampus to brain regions responsible for emotional response may better enable these central hubs of information processing to direct learning towards phenomena which are most emotionally salient. This appears to be the approach the brain takes, as learning and memory have long been understood to have a strong, if complex, connection to emotion (Tyng et al. 2017). This may better enable the brain to focus learning on patterns which are likely to lead to positive or negative outcomes.

Moreover, by connecting related entities by means of a central processing hub, the brain may also exploit a form of ensemble learning; rather than allowing related entities to directly connect, enforcing some separation between related entities which are composed into a single representation elsewhere may better enable the formation of independent lines of evidence for the same latent entity. This in turn may also better enable richer composite representations which allow for more independent, context-dependent representations.

Following from this discussion, the relative merits of localized connection growth compared to a global approach remains an interesting open question. However, it is also important to emphasize that connection growth of neurons in the brain does not require the proximity of the cell bodies of neurons, only proximity of the axons of presynaptic neurons to the dendrites of postsynaptic neurons. Whether this localized growth strategy is essential to brain function remains an open question. How such a growth strategy which considers the distance between axons and dendrites can be implemented efficiently also remains an open question.

\subsection{Emergent Behavior of Livewired Neural Networks}
There are a broad variety of behaviors in humans which can be more clearly understood by assuming a livewired view of learning that introduces connections to represent new relationships between entities.

The most illustrative example of the explanatory power of livewiring can be seen in the case of sensory substitution, wherein the functional connections between brain regions change to maximize access to rich sources of information about the external environment. For instance, blind people can learn to process visual information using electrodes implanted on the tongue that stimulate patterns captured by a camera (Bach-Y-Rita \& Kercel 2003). 

More broadly, in individuals with sensory deficits, a variety of functional changes in brain structure have been observed that enable the substitution of sensory modalities to the extent to which such information is available from the environment (Merabet \& Pascual-Leone 2010). Under a livewired model of the brain, such changes to the functional structure of the brain would be enabled by substantial connection growth between brain regions which can derive predictive value from mutual communication.

Another phenomenon which can be clearly explained by livewiring is that humans have a bias toward inferring causal connections between pairs of events that happen to coincide (Foster \& Kokko 2009, Johnson et al. 2013). In this case, under a livewired interpretation, the network regions which represent the most prominent entities involved in each event will be strongly activated coincidentally, giving rise to new connections between these entities, enabling learning of potential causal connections between coincidental events.

However, it should be noted that new connections between previously unrelated phenomena can be erroneous in the case of apophenia (Conrad 1958), suggesting that newly formed connections should remain open to change to ensure spurious relationships are not consolidated too quickly, following from the credible credit assignment reasoning described above. 

For another clear example of how a livewired interpretation can help explain emergent behaviors of brains, it helps to reflect on the size of brain regions devoted to processing touch sensory information from different body regions. In particular, it has long been known that the size of brain regions devoted to touch does not map to the physical size of each corresponding body part, but rather roughly corresponds to the amount of information each area provides to the brain. This gives rise to brain regions with a distorted representation called a “homunculus” that devotes more resources to sensitive areas such as lips and hands and far less to less sensitive areas such as the torso (Penfield \& Boldrey 1937). 

Under a livewired interpretation of brain activity, this distorted map would naturally form in response to stimuli from the environment; sensory systems which offer the most salient details of the environment will frequently provide the brain with useful information, generating strong activations which are reinforced, and causing surrounding neurons to rewire to connect to this rich source of information. In this way, sensory modalities which are most informative for future experience will expand their influence to larger network regions, giving rise to the type of distorted map presented in the homunculus. 

Conversely, sensory modalities which do not provide useful information about the environment will have correspondingly small brain regions devoted to processing them. This can be observed in the ability of occipital lobes of blind individuals to be recruited to process auditory information rather than visual. However, it should be noted that the extent of this role reassignment is related to the age at which the individual was first blind, supporting the idea that the learning rate and rewiring rate for the brain decreases with age (Collignon et al. 2013).

A further example of the explanatory power of a livewired model of brain function can be seen by considering the mnemonic device of the method of loci, otherwise known as memory palaces. To summarize, the method of loci involves memorizing information by first picturing a well known place in the mind’s eye, then populating that space with objects or people meant to correspond to the items to be memorized. This approach is most effective when picturing objects with emotional salience (Qureshi et al. 2014).

Under a livewired model of the brain, the success of this approach can be clearly understood; the brain region(s) devoted to representations of highly familiar locations and familiar objects or people should be expected to be quite large, so there are many neurons to which new connections can be made, enabling far more opportunities for the brain to find pointers back to the original item to be memorized. Put simply, the memory can be stored across a broader region of the brain by connecting to familiar elements. 

In a similar vein, livewiring can also offer an explanation of the power of analogies to explain complex topics; concepts which are familiar and commonly experienced should be expected to have a significant amount of representation in a livewired brain, so by connecting the structure of a novel idea to an analogous idea which is well understood, one can exploit a larger brain network to process the new information in a structure-preserving way.

A common description of these prior examples is that the brain organizes itself around rich sources of information, a tendency termed infotropism (Eagleman 2020). As seen in these examples, rich sources of information need not be directly from the current environment; rather, they often come from latent representations based upon prior experience.

In forging connections between higher-order representations of the environment, we also begin to see a potential neuronal basis of priming, whereby objects or ideas prime us for recognizing related objects or ideas (Foss 1982). Namely, related objects and ideas are expected under a livewired interpretation to have connections between the network regions used to represent them, so activation in one area will prepare the brain for activation in connected areas.

For further examples of emergent brain behaviors that can better be understood by applying a livewired view of network structure, see Eagleman (2020).

\section{Areas for Further Development}
While livewiring is a promising approach to achieving more compact, relational models of the world which are capable of stable few-shot learning, it also raises numerous questions that are deserving of additional focus beyond the scope of this work. I will explore some of these below, including questions regarding how livewired neural networks could explain phenomena in both neuroscience and psychology. 

As it has long been thought that the connection structure of brains is in large part responsible for the behaviors brains generate, it should be expected that a model of brain behavior that predicts dynamic connection structure should generate substantially different predictions about the emergent properties of brains than a model that assumes a fixed structure. I will attempt to articulate a number of the most significant of these potential lines of inquiry below.

\subsection{Foundational Questions}
Effectively fitting neural networks is often an iterative process, attempting to fit models using different architectures and hyperparameters, applying heuristics to guide this process. It seems reasonable that such optimizations should also apply to livewired neural networks.

For instance, it has been experimentally demonstrated that the initial structure of neural networks can strongly impact their ability to train effectively. More specifically, network depth has been shown to be a powerful tool for building more capable models (Szegedy et al. 2014), but simultaneously, shallow effective network depth, which can be achieved by using connections across multiple layers, has been shown to be important to performance (Larsson et al. 2017). Presumably, the benefits of these characteristics should extend to livewired neural networks. 

However, different initialization structures may be more beneficial to livewired networks, which should be investigated. Here, I have suggested the use of a fractal pattern of exponentially increasing sparsity when spanning multiple layers under the assumption that it should efficiently induce a sparse network structure while minimizing effective network depth. However, the choice of branching factor may impact the performance of the network. The optimal branching factor may even vary depending on the region of the network.

To illustrate this point, it helps to consider the density of synaptic connections in different brain regions throughout the aging process. It has been demonstrated that some brain regions form more connections at a young age, yet other regions grow connections more quickly (Huttenlocher \& Dabholkar 1997). More specifically, sensory regions of the brain form earlier, with the forebrain taking longer to fully develop (Gogtay et al. 2004).

However, in all brain regions, following the initial growth stage, connections are pruned as brains age, moving the brain toward greater sparsity (Tierney \& Nelson 2009). This strategy is consistent with the Lottery Ticket Hypothesis, which suggests that only a sub-network of a randomly initialized network is needed for learning, but that the presence of many ultimately redundant connections increases the likelihood of finding a successful “lottery ticket” which quickly converges to a good solution (Frankle \& Carbin 2018). 

While measurements of synaptic density that differ across brain regions do not directly reflect the prevalence of long range connections, they do at least suggest 1) different regions of neural networks should be initialized with different levels of sparsity, possibly including different ratios of long-range to short-range connections, and 2) different network regions should grow new connections at different rates, possibly in response to the performance of that region.

To expand upon this last point, an important question is how the density of livewired networks should be adjusted over time to optimize the resources consumed in the learning process. Using the connection formation strategy of the brain as a guide, it would seem that network areas closer to sensory inputs should converge more quickly and thus have a shorter period of net positive connection growth, while higher-order network regions should maintain net connection growth for longer. 

Ideally, connection growth would not be determined by a set schedule, but rather activity-dependent in each region in response to the quality of the predictions that region forms. This is desirable because continuing to grow new connections beyond the point of convergence to a successful representation is at best a waste of computational (or metabolic) resources, and may also introduce needless noise to a good model. 

However, how to achieve such experience-based connection growth rules remains an open question. It is possible that in the brain, there is no learning-based strategy for this approach, but rather that growth rates are genetically predetermined, which appears especially likely for prenatal brain development in the absence of much meaningful stimuli.

A more specific question of interest for livewired networks as described above is how many new connections should be allowed to grow at each step? Presumably, the more connections are allowed to form (with other connections pruned to keep the sparsity balanced), the more respondent to change the network will become. In this way, the rate of connection formation can be likened to the role of the learning rate, suggesting it should decrease over time to enable rapid improvement in early stages when the model is not expected to perform well, but to remain more stable in later stages to enable model convergence.

However, it has been demonstrated that a strictly decreasing learning rate schedule is not necessarily optimal for convergence to a strong solution, but rather setting the learning rate to increase initially before decreasing can accelerate model convergence (Smith 2017, Smith \& Topin 2017). As argued above, this can be linked to the problem of credible credit assignment. Given this comparison, it appears reasonable to conclude that the rate of new connection formation should also be responsive to recent values of the loss function. 

In particular, similar to the learning rate, it seems reasonable that the rate of connection formation should increase while the loss function remains high, but should decrease as the loss decreases to enable convergence to a stable solution. While this approach appears consistent with past findings, it should be confirmed experimentally, including testing of different connection growth rate update rules.

Additionally, astrocytes in some areas of the hippocampus exhibit more active calcium variations than others (Fields et al. 2015). This further points to the possibility that there may be value in varying the learning rate and the connection formation rate across network regions. Credibility may serve a useful role in guiding this idea further.

\subsection{Alternative Livewired Approaches}
While I have focused on growing new connections between hidden nodes, another valid approach would be to also form new neurons, possibly by copying nodes that have many outgoing connections, with a random subset of the existing connections on that node and random weights (or random perturbations of existing weights on that node). 

This is a reasonable approach because when a neuron has many outgoing connections with large weights, that indicates that the representation that node forms is highly informative. Thus, we should enable more processing power to be devoted to understanding what it represents, and larger networks can produce more sophisticated representations. To further justify this approach, it should be noted that such neurogenesis is implemented in some brain regions (Bergmann et al. 2015).

An additional potential benefit of allowing the growth of new neurons is that an existing model could easily be transferred to a machine with greater resources, maintaining the existing learning while enabling the expansion of the model to further exploit the available computing power.

To see how else this process of node generation might be useful, it helps to reflect on the size of brain regions devoted to processing touch sensory information from different body regions. In particular, recall the distorted representation of the homunculus within the brain that devotes more resources to sensitive areas such as lips and hands and far less to less sensitive areas such as the torso (Penfield \& Boldrey 1937). 

By providing additional neurons to network regions responsible for processing especially salient information, as measured by the weight of connections emanating from particular network regions, richer representations may be formed of entities which have proven to be significant in forming accurate representations of the environment.

One potential problem this approach could help to overcome by selectively growing network regions is that brains have had the benefit of millions of years of evolution to design the initial size of different network regions, while artificial neural networks must be initialized to a particular size. However, the growth of new neurons will likely require the pruning of other connections or neurons to remain within the computational constraints of the entire network. 

One possible objection to this neuron growth approach is that while the brain does grow new neurons through neurogenesis (Ming \& Song 2011), neurons are often reallocated from neighboring brain regions to expand computational capacity for tasks that are important (Pascual-Leone et al. 2005), which would call into question the necessity of growing new neurons. 

However, while in biologic systems, the number of neurons which can be accommodated is limited by the metabolic constraints of the body (and the size of the skull), this is a much looser constraint in artificial neural networks due to continually increasing hardware capabilities. Once again, while initial network designs may be constrained by current hardware limitations, enabling the growth of a larger, more powerful model on top of an existing one is a promising approach for building models with continually increasing representational power.

One empirical question of interest for livewired neural networks is whether the locality of new connection formation in the brain is actually a useful feature in some contexts. It is possible that livewiring should exploit all coincident activations in disparate nodes, interpreting all such coincidences as evidence for conditional relationships between the respective features. In this case, the locality of neurite outgrowth in the brain may solely be due to physical limitations imposed by the metabolic and physical constraints of the brain. 

However, it is also possible that there are other emergent properties of the localized growth patterns of the brain that may provide benefits under some circumstances. For instance, such a constraint may encourage network modularity, maintaining largely separate systems for distinct tasks, which would explain the separation of brain regions by function.

One further question which deserves further investigation is whether alternative connection growth rules are more efficient at generating relational representations. In particular, rather than considering only forming connections between pairs of neurons with strong activation values, perhaps models that consider forming connections between the direct ancestor or child nodes of neurons with strong activations would learn better representations. 

To understand why such indirect connections may provide some benefits in comparison to forming direct connections between related entities, recall that for priming in humans (Foss 1982), there is no sense of directionality in the associations between entities. However, by drawing a connection from an earlier layer to a later layer, the bi-directionality of associations between entities may be somewhat limited. In this case, it may be more efficient to consider connections to the child nodes of the node in the later layer, rather than direct connections. The comparative efficiency of these two approaches is an open question. It should also be noted that attention-based methods may provide a means to introduce bi-directionality to the network.

By focusing connection growth on child nodes, models may also be better able to form compositional concepts which combine multiple entities into a new, separate entity. Such an indirect rewiring approach may also better reflect the locality of connection growth in the brain. As suggested above, while such locality may largely be a consequence of the metabolic and size constraints of the brain which would preclude a global signaling mechanism across brain regions, it is also possible that this locality better enables the brain to maintain functional separations between brain regions, in which case indirect connection approaches may prove more valuable.

If indirect connections to parent or children nodes are considered, one further modification which may be valuable is bundle rewiring, whereby connections to multiple nodes may be inspected for formation. This approach may especially be valuable for forming indirect connections between nodes in the same layer which coincide.

One further modification to the livewired approach which is deserving of further investigation is the use of recurrent connections between individual nodes to enable the growth of backwards connections across time periods. This could serve as yet another means for features from later layers to influence features from prior layers without interfering with the use of backpropagation. The relative merit of this approach in comparison to attention-like approaches or the use of indirect connection approaches is an open question.

One further question of empirical interest is whether the individual neuron is the appropriate level of abstraction at which to characterize an entity. To this end, it is important to note that the cortex of the brain is arranged in columns of six distinct layers (Hawkins et al. 2017), hinting at the possibility that a small cohort of neurons, rather than individual neurons, may provide the appropriate level of detail at which livewiring should occur.

Finally, while I have focused here on the application of livewiring to more general neural networks, there may be some additional benefits to applying livewiring to graph neural networks. In particular, graph neural networks may be better able to exploit known relational structures in the environment to initialize structures atop which further livewired connections may be built.

\subsection{Sparse Activations}
At any given time, only a fraction of brain regions are active (Kerr et al. 2005). This sparsity is likely an important feature which limits the metabolic consumption of brains. If this approach could be replicated in neural networks, it could potentially enable greater efficiency by eliminating the need to calculate connections which are not likely to significantly impact the calculations of the network.

While dropout can be thought to roughly mimic this sparse activation process by randomly selecting connections to drop (Hinton et al. 2012), the availability of activation statistics provided by a livewired approach hints at a more principled method to dropping connections: drop outgoing connections from nodes with low activation values, either all connections, or in a probabilistic way that depends on the activation value. 

Such an approach might better reflect the all-or-nothing firing approach of neurons in the brain, and may reduce the runtime required for the network without significantly impacting its performance. However, this point warrants further investigation.

While this sparse activation approach is promising for discriminative models, possibly enabling them to calculate nearly the same functions with fewer resources, it is perhaps more promising for generative models which enable models of the future to anticipate how the world may evolve. In particular, in predicting future world states, such projections should likely only entail a small subset of the entire possible latent space. Given the large branching factors of real environments, sparsity is likely critical to forming predictions at a reasonable level of detail.

To further illustrate the priority the brain places on the efficiency of its representations, note that it has been observed that experts (at least in certain task environments) use less of their brain when completing a familiar task than do novices when completing the same task (Haier et al. 1992). Reflecting on the interpretation of the brain as minimizing the need for environmental information, this emphasis on efficiency becomes more clear; when first learning a new task, the representation of the given environment should be expected to be inefficient. The approach the brain appears to adopt seems to follow the approach suggested by the Lottery Ticket Hypothesis (Frankle \& Carbin 2018); initially devote a large brain region to a new problem and update connection strengths to achieve a better representation. 

However, as learning proceeds, the most efficient pathways are retained, becoming hard-coded for the desired behaviors, while pathways not part of the winning ticket are reallocated to other representations. In particular, the pathways of the brain are held in constant tension by the rewiring process, opening the possibility of any redundant pathways being pruned to make way for new representations. In this way, livewiring enables the brain to maximally exploit rich new sources of information by expanding the pathways of representations which prove significant, a pattern that has been termed infotropism (Eagleman 2020).

Connecting this drive toward efficiency back to the notion of dropout, an additional potential benefit of activation-driven dropout is that it will intentionally restrict the portion of the network devoted to solving a particular task. Tying the extent of the dropout to the performance of the model on a given task may improve the efficiency of this process, though how best to achieve this is an interesting open question.

Moreover, by restricting the portions of a network involved in a particular learning task, this activation-driven dropout approach may also mitigate updates to parts of the network which provide comparatively weak learning, thereby minimizing the noise introduced to network regions largely unrelated to the current task. Once again, this benefit of model stability should be balanced against the performance of the model.

Alternatively, there may also be benefits to dropping connections from nodes which have recently experienced strong activations. In the brain, some axons have the ability to blockade certain branches, impeding the transmission of action potentials. This occurs most often in the largest branches of axons following a period of high frequency stimulation (Debanne 2004).

While this approach may appear counter to the notion of sparse activation, recall that the motivation behind dropout is to ensure the development of redundant pathways by preventing the co-adaptation of feature detectors (Hinton et al. 2012). Dropping signals from nodes expected to have the strongest activations may thus be interpreted as a means to ensure nodes which are most likely to contribute to co-adaptation of features are limited in their ability to do so.

\subsection{Long-Term Potentiation}
In brains, long-term potentiation (LTP) primes neurons with strong activations for additional subsequent activations, lasting as long as days (Nicoll 2017).

To understand why this may be desirable, recall that the representations formed by neurons can be interpreted as higher-level entities in the environment. Operating under the assumption that the environment is reasonably stable, then, we can see that introducing LTP will enable neural networks to better model environments which demonstrate object permanence, or state persistence more broadly. 

In particular, in an environment in which entities persist through time and space, the best guide to entities expected to be encountered in immediate future experience is recent past experience. Thus, by priming the model to recognize entities that were recently encountered, LTP should enable a form of distributed short-term memory across the entire network.

However, one possible concern this approach raises is that maintaining separate potentiation states for each neuron or connection would impose a somewhat large memory cost. Indeed, merely expanding the memory allocated to attention-based mechanisms may be more efficient in artificial neural networks, as the massive parallelism of the brain is computationally expensive to replicate. This tradeoff is an interesting open question which will likely require robust experimentation across a range of tasks.

Regardless of the relative merit of LTP in comparison to attention-based methods, enabling the persistent representations of states will likely be a critical component of enabling livewired systems to learn the most important relationships in the environment, as such relationships are typically not precisely contemporaneous, but rather spread across time. More practically, credit assignment should focus on updating the relationships with entities that have been recently encountered to form better representations. 

In terms of how such LTP may be implemented in artificial neural networks, there are at least three potential approaches: 1) apply factors to the weights of connections based on LTP, though this may be challenging, as the weights of the model have been optimized by the learning process, 2) apply LTP factors to activation values, though once again this may interfere with prior learning, or 3) enable LTP factors to influence the dropout behavior of sparsely activated networks. However, the relative benefits of these approaches is an open problem.

\subsection{Role of Neuromodulators}
Given the profound, complex relationship between emotion and learning (Tyng et al. 2017) and the premise that livewiring is a critical aspect of the learning process, it is natural to ask how this influence is achieved, and to what extent the livewiring process might be influenced by emotional states.

One of the most influential mediators between emotions and learning are the dopaminergic pathways, which play a significant role in learning. Dopamine is strongly tied to feelings of pleasure and reward, and plays a significant role in shaping goal-directed behavior (Arias-Carrion et al. 2010). Dopamine is primarily produced in only a few brain regions, from which it is transported broadly across other brain regions by dopaminergic neurons whose axons project to these regions through the mesocortical, mesolimbic, and nigrostriatal pathways (Prasad \& Pasterkamp 2009).

Given dopamine’s apparent significance to the learning process, we would expect under a livewired interpretation of learning that relies on astrocytes to guide connection growth that the presence of dopamine would stimulate the types of morphological changes in astrocytes we have come to expect as part of the livewiring process.

In fact, dopamine has been found to initiate large intracellular calcium increases in cultured hippocampal astrocytes (Parpura \& Haydon 2000). Moreover, extracellular dopamine has been found to induce profound morphological changes in astrocytes, causing them to stellate more new processes as dopamine concentrations increase, enabled by astrocytes’ dopamine receptors (Galloway et al. 2018).

Along a similar line, it should also be noted that stress leads to release of glucocorticoids (Jauregui-Huerta et al. 2010), and that glucocorticoid receptors in astrocytes have been shown to be important to the formation of aversive memories (Tertil et al. 2018). 

Given the above, it would seem that neuromodulators released due to emotional responses significantly impact the livewiring process by altering the behavior of astrocytes. Given the broad influence of the dopaminergic pathways across much of the brain, one could argue that dopamine may serve as a global signaling mechanism across the brain that influences the connection formation rate in response to reward signals from the environment. This increase in network plasticity gives rise to more substantial changes to network structure, thus enabling more substantial model changes in response to reward signals.

Additionally, it has been observed that a number of neuromodulators are disseminated throughout the brain through cerebrospinal fluid in the ventricles (Veening \& Barendregt 2010). This may best enable the influence of neuromodulators that need only operate on longer time scales; for instance, it has been hypothesized that the ventricles disseminate neuromodulators that regulate sleep and appetite (Tan et al. 2010, Zappaterra \& Lehtinen 2012). The impact of these on the behavior of glial cells in altering the network structure of the brain is an interesting area for further study.

\subsection{Extensions of Few-Shot Learning}
One empirical question of interest is whether it would be more efficient to conduct gradient-free connection growth, where any connections between nodes with large activation values are formed. This would eliminate the need for computationally expensive gradient calculations. Given the focus on nodes with large activation values, this approach is still likely to generate connections with large loss gradients due to the fact that the loss gradient for a connection is proportional to the activation value of the parent node (Rumelhart et al. 1985).

One possible approach in which this gradient-free connection scheme may be especially beneficial is to interleave stochastic gradient descent and the growth of new connections. In particular, it may be valuable to iteratively perform stochastic gradient descent and rewiring, with the number of iterations contingent upon the magnitude of the loss function. 

Regardless of the benefit of applying the loss gradient to selecting growth between strongly activated neurons, this iterative approach to rewiring and fitting new connections is a promising approach to achieving efficient few-shot learning. Using this approach, a model would devote more time and resources to understanding situations for which its existing model is poor. This can be directly compared to humans’ tendency toward curiosity, whereby we pay most attention to aspects of the environment which we do not understand.

By focusing learning on items for which the model performs poorly, and by layering small model  modifications in response (in the form of new connections), this approach can be compared to gradient boosting machines (for an overview, see Natekin \& Knoll, 2013). Moreover, given the focus of livewiring on nodes for which the presence of a higher-order entity is strongly indicated by the environment, this error refinement approach may focus on developing richer representations at a symbolic level. This point remains open to further investigation.

\subsection{Foundational Neuroscience}
While livewiring presents an explanation for how the brain may solve the binding problem, offering an explanatory framework for both the low-level details of the roles of glial cells in managing the connections between neurons and the high-level observations of associative emergent brain behaviors, there remain further predictions of livewiring which can be validated experimentally.

Returning to the illustrative case of sensory substitution, it has already been observed that the functional changes in the brain induced by this process are enabled by significant updates to the connections between brain regions at larger scales (Bach-Y-Rita \& Kercel 2003). However, a livewired model of the brain would predict more specifically that this large-scale update to the functional map of brain regions would be enabled specifically by substantial changes to the connection structure of individual neurons. 

More specifically, during periods of substantial changes in functional structure such as during a period of learning to perform sensory substitution, astrocytic calcium waves above baseline levels may be expected in the involved brain regions, increasing the motility of astrocytic processes in those brain regions, giving rise to the growth of axonal pathways between the brain regions which need to communicate.

However, how this growth of new connections proceeds exactly is an interesting open question. For instance, as discussed above, the brain makes use of centralized hubs, such as the hippocampus and prefrontal cortex, to consolidate information; it is possible, then, that the livewiring process could be initiated in such regions, given the inherently local nature of livewiring in the brain. 

However, it is also possible that livewiring could be a more distributed process. To support this idea, it should be noted that brain regions associated with advanced cognitive functions, such as the cortex and the hippocampus, contain pyramidal neurons with many branching processes (Spruston 2008), which could help facilitate livewiring in a more distributed fashion by more widely disseminating the information provided by strong neuron activations in the cortex.

Given the previously discussed locality of astrocytes in directing the growth of such new connections, a more specific prediction that would also arise is that the rate at which such connections form should have an inverse relationship with the physical distances between the two brain regions. However, the complex structure of connections in the brain could complicate this, as the prior presence of long-range connections between regions could increase the rate of new connection formation.

Another specific prediction livewiring as presented above would offer is that given that it has been observed that intercellular calcium waves in astrocytes spread to twice as many cortical and hippocampal astrocytes as between astrocytes from the hypothalamus and the brain stem (Blomstrand et al. 1999), and the corresponding increase in astrocyte motility in response to calcium elevations (Bernardinelli et al. 2014), the hippocampus and cortex should be more variable in their connection structure comparatively.

It has long been known that sleep is vital to memory consolidation (Rasch \& Born 2013). Applying a livewired model of the brain to changes the brain undergoes during sleep offers very specific interpretations of the mechanisms that may in part be responsible for this relationship.

For example, microglia eliminate synapses during sleep, and molecules that signal synapses should be eliminated are increased during sleep (Choudhury et al. 2019). Under a livewired interpretation, the need for this process to occur during sleep becomes more clear; subjecting the conscious brain to eliminations of its existing connection structure could interfere with conscious thought processes, but this constraint is relaxed in the sleeping state. Indeed, the often disordered nature of dreams could be a reflection of the impact this connection pruning process has on subjective experience.

One important question deserving of further consideration is how microglia select connections to prune. In particular, it may be useful to consider whether there are global signaling mechanisms microglia make use of to determine how many connections to clear, or whether this process is entirely local, and if so how such connections are identified. In artificial neural networks, I have argued that connections with weak weights should be pruned first, though whether there is an analog which microglia encounter is an important question to answer clearly.

Additionally, it has been observed that astroglial processes retract during sleep, retreating from their coverage of synapses as occurs in waking states (Bellesi et al. 2015). Apart from increasing the exposure of synapses to enable microglia to perform their pruning activities, under the interpretation offered by livewiring, this retreat also suggests that the formation of new connections is restricted in sleep. 

To understand why this should be expected in a livewired model, recall the purpose of changes to the connection structure of neurons: to connect related entities. As suggested above, unconscious brain activity during sleep is not tied to direct sensory experience, and so can become disordered. 

If the brain were to attempt to remember these experiences, then, it may generate connections between entities that were only coincidentally activated by chance rather than due to input from the external environment. This could generate random connections with little predictive power of the external environment, and so it is reasonable for the brain to de-prioritize the development of new connections during sleep to prevent the formation of such spurious connections. This would explain why dreams are difficult to remember; we do not lay down new pathways during sleep to recall such experiences.

\subsection{Livewiring and Brain Dysfunction}
Assuming livewiring mediated by glial cells plays a significant role in the development process of the brain, one would expect that brain disorders would in many cases be related to dysfunctions in the glial cells which help regulate the brain’s network structure. This expectation matches observations of brains across a wide variety of disorders, and efforts to better understand the role of glial dysfunction in disease remains an active area of research (Barres 2008). 

For instance, Alzheimer’s has been found to be associated with the dysfunction of microglia, possibly involving them eliminating synapses too aggressively (Mosher \& Wyss-Coray 2014). Similarly, overactive pruning of synapses by microglia and astrocytes have been implicated in Parkinson’s disease (Tremblay et al. 2019). More generally, dysfunction of glial cells, particularly astrocytes, has been found to be associated with a broad variety of neurodegenerative disorders (Oksanen et al. 2019).

The end result of many of these varied disorders is a substantial alteration to the structure of the connections between neurons, typically involving the destruction of many of the prior connections. This reinforces the importance of the network structure of the brain in determining its function. 

However, in exploring brain pathologies, potential dysfunctions under distinct failure modes are not the only source of insight into brain function; effective treatments can also offer clarity into the core mechanisms upon which the brain relies. 

For instance, it has been found that there is a great deal of variability in the severity of symptoms experienced by patients with neuronal hypertrophy consistent with Alzheimer’s; patients with substantial brain lesions who lead a more active lifestyle tend to develop less severe symptoms. This has led to speculation that despite the loss of connections caused by disease, the brain remains able to form new connections to replace the lost ones, thereby maintaining greater functionality (Iacono et al. 2009).

In the case of severe depression, electroconvulsive therapy, one of the most effective treatments for severe cases, has been shown to generate increased dendritic arborization in the dentate gyrus (thought to be responsible for encoding episodic memories) in animals, and has been shown to update the functional connections from this region in humans (Takamiya et al. 2019). 

From all of these cases of brain dysfunction and crude remedies, we see that alterations to the livewiring process have a significant impact on brain behavior. This further reinforces the importance of focusing future research on ways in which the behavior of glial cells can be better regulated to maintain healthy brain functionality.

\subsection{Self-Supervised Learning, Efficient Encoding, and Transfer Learning}
Assuming livewiring networks give rise to operations on symbols as argued above, it is natural to conclude that such models will achieve far more compact representations of the environment, especially environments that evolve over time, as operations on symbols allow for far more efficient transition models for environments with stable entities (see Yi et al. 2018 for an example of such operations on manually defined entities). More efficient world models should naturally lead to an improved ability to efficiently encode the most important information in the environment to form predictions about the future.

Any improvements in the efficiency with which livewired networks represent the environment should generate corresponding efficiency gains in self-supervised models which attempt to reason about future world states on the basis of past observations. In particular, performing operations on efficiently encoded symbol-like entities should greatly reduce the space of future world states which must be examined. 

To emphasize the importance of the efficiency of representations to enabling effective inference, it should be noted that even in humans, the ability to retain information in working memory is constrained by the efficiency with which information is organized. Organizing such items in working memory efficiently is referred to as “chunking” (Thalmann et al. 2019).

Moreover, the emphasis livewired networks place on learning latent representations of recurring entities in the environment may also enable such models to achieve more efficient multimodal learning (Ngiam et al. 2011) across different domains. In particular, as the same latent entities in the environment should persist across sensory modalities (to the extent to which such entities can be sensed by separate modalities), information from one modality should enable the identification of latent entities and thus the projection of future experience of other sensory modalities. Additionally, a more symbol-driven learning system may enable greater context dependence by identifying latent entities which influence the interpretation of sensory inputs.

Perhaps the greatest tool humans use to organize multimodal learning is language; language by its construction enables humans to develop identical latent representations of the environment, and to share such representations with others without the need for directly shared sensory experience. Moreover, as language is designed to describe the entities in the environment which are most relevant to human experience, it provides a powerful organizing tool around which people can structure their experiences. Indeed, translation is only possible because all languages take advantage of approximately the same structures in the environment.

Any human-level learning agent which has been exposed to language can thus be expected to develop a world model which is built around the same language structure as speakers of that language, organizing internal representations around the same latent entities as suggested by the language. The importance of exposure to language can be seen in the impaired development of individuals who are not exposed to language during early development when the brain is most plastic (Hall 2017).

Language also provides a tool by which novel experiences can be characterized in terms of existing knowledge. In particular, learning agents which operate on symbols should be capable of applying analogies, and language provides a useful level of abstraction at which this process can be accomplished. This should greatly facilitate the transfer of learning between problem domains.

Attention models have had great success in advancing state-of-the-art performance on tasks requiring semantic understanding (Vaswani et al. 2017, Chaudhari et al. 2019). Assuming livewired networks greatly improve the efficiency of information retention in such systems, such models should be expected to become yet more effective.

\subsection{Attention Models and Online Self-Supervised Learning}
The persistence of calcium waves in astrocytes for 5-30 min. (Scemes \& Giaume 2006) suggests an agent pursuing online learning should maintain the strongest activations in individual nodes in the priority queue for future connection growth well after the initial activation. To understand why this may be useful, it helps to consider that in operant conditioning, whereby animals can be taught to associate behavior in response to stimuli with rewards (or punishment), the schedule of rewards following an operant is highly significant to learning (Staddon \& Cerutti 2003), enabling stimuli which are temporally close to be recognized as related, with the strength of the relation dependent upon the time elapsed.

Analogously, in the case of online learning, it will likely be beneficial to maintain nodes with especially strong activations in the priority queue of nodes inspected for new connections. To replicate the temporal dynamics suggested by operant conditioning, it may suffice to apply a constant forgetting factor hyperparameter to all prior activations. 

Alternatively, the persistence of past items for learning could be achieved by relying on attention mechanisms to carry over information from past periods, or by using long-term potentiation to amplify the activations of nodes recently activated strongly. An attention-based approach may better enable reward information to guide the persistence of the most important information, though a combination of these approaches may best enable the persistence of different types of information.

Regardless of the specific mechanism(s) used to achieve learning across time periods, one constant across these approaches is that learning should be strongly linked to an agent’s expectation of reward; in humans, learning is strongly (if complexly) related to emotional state (Tyng et al. 2017).

To roughly characterize how emotions and rewards should drive the retention of information across time to enable self-supervised learning, there are a few forces which should regulate learning: 1) curiosity, which directs agents to retain more information concerning novel domains to learn about potential sources of rewards or punishments, 2) fear response to cases of punishment, and 3) anticipatory response to expectation of rewards.

In each of these cases, the time spent reflecting on experiences, and the extent to which information about such experiences is retained across time, should be driven by the anticipated rewards or punishment associated with the experience. For cases of greater reward signal uncertainty, the drive towards curiosity should incentivize even more resources to be devoted to learning. This approach is consistent with observations that uncertain rewards generate even stronger dopamine responses in humans than certain ones (Niv et al. 2005).

To enable this type of behavior, extensive connections should be established initially between network regions associated with reward and punishment expectation, attention, and planning given the functional importance of these links. This is consistent with the architecture of the human brain, where components associated with the anticipation of rewards and punishments, particularly the dopaminergic pathways and the limbic system, are tightly linked to brain regions largely responsible for working memory and planning (Arias-Carrion et al. 2010, Messe et al. 2014).

\subsection{Generative Models and Connecting the Dots}
As discussed above, learning in a livewired network can be compared to the search for novel connections between existing symbols. However, the power of such an approach will not be limited to discriminative models; generative models should be expected to be able to make use of the interactions between symbols to form new representations using existing symbols.

In particular, generative models already allow for style transfer between images (Karras et al. 2019); analogous methods may allow for applying newly learned knowledge to an existing knowledge base to generate new knowledge, relying on symbolic structure to follow a chain of reasoning. 

Enabling reward signals to guide the time and resources devoted to the creation of such novel representations is a promising approach, which would be consistent with the significance of emotional salience to learning (Tyng et al. 2017).

To provide additional context for such a potential iterative process of learning and inference, it should be noted that during periods of rest, including sleep, the hippocampus replays recent experiences, which is generally acknowledged to be critical for memory consolidation and learning (Olafsdottir et al. 2018).

\subsection{Unpacking the Black Box}
Assuming livewired neural networks enable relational models to operate on symbols, we can begin to see how they might be used to build models of the world that more directly reflect human thought processes. In particular, by forming associative representations of entities in the environment, livewired networks may be capable of forming the same relations as humans find useful in describing the world.

For instance, livewired networks may be able to use their ability to manipulate symbols to use language as a model of the world. This in turn may enable livewired networks to connect entities in the environment with the words used to describe them. This may enable livewired networks to perform highly efficient transfer learning by exploiting the learned structure of language to form a model of the aspects of the environment which language describes. 

Moreover, by enabling models to manipulate symbols, especially language, which correspond directly to internal representations of entities in the environment, livewired networks may greatly advance the interpretability of such models. This in turn could make eliminating bias from a model far more transparent by directly investigating the associations formed by the model and increasing the ability of such models to generate explanations for their behavior.

More specifically, by investigating the associational structure of the entities represented by a livewired network, a clear interpretation of the biases of the network could be obtained. To overcome undesirable representations, training data which explicitly contradicts the objectionable views could be repeated to the model until its representation matches the desired behavior. To ensure the model is receptive to updating its prior beliefs, the learning rate and connection formation rate could be temporarily increased to encourage the model to keep an open mind.

As with human learners, the representations formed by models will come to reflect the information that is available to them, and which is incentivized by reward signals. Great care must then be taken to ensure that information is presented with appropriate context, and that behavior is carefully monitored to ensure the proper actions are incentivized.

\section{Conclusion}
Here I have illustrated how livewired neural networks may be expected to give rise to models which operate on symbol-like entities. I showed how the growth of new connections between existing neurons, paired with a higher learning rate for newer connections, may enable such livewired networks to achieve few-shot learning without overwriting previous learning. 

Furthermore, I have demonstrated how this livewired approach could give rise to a number of emergent behaviors in humans, and provided a plausible biological mechanism by which such livewiring may be accomplished in the brain.

Finally, I have provided a number of promising directions for further research pursuing the premise that livewiring presents a substantially novel model of how the brain adapts to the environment.

\bibliographystyle{unsrt}

\end{document}